Article

# Neural Network Entropy (NNetEn): Entropy-Based EEG Signal and Chaotic Time Series Classification, Python Package for NNetEn Calculation

**Andrei Velichko** [1,*], **Maksim Belyaev** [1], **Yuriy Izotov** [1], **Murugappan Murugappan** [2,3,4] **and Hanif Heidari** [5]

[1] Institute of Physics and Technology, Petrozavodsk State University, 185910 Petrozavodsk, Russia; biomax89@yandex.ru (M.B.); izotov93@yandex.ru (Y.I.)
[2] Intelligent Signal Processing (ISP) Research Lab, Department of Electronics and Communication Engineering, Kuwait College of Science and Technology, Block 4, 13133, Doha, Kuwait; m.murugappan@kcst.edu.kw
[3] Department of Electronics and Communication Engineering, Faculty of Engineering, Vels Institute of Sciences, Technology, and Advanced Studies, 600117, Chennai, India
[6] Centre of Excellence for Unmanned Aerial Systems (CoEUAS), Universiti Malaysia Perlis, Arau 02600, Perlis, Malaysia.
[5] Department of Applied Mathematics, Damghan University, Damghan, Iran; heidari@du.ac.ir
* Correspondence: velichko@petrsu.ru

**Abstract:** Entropy measures are effective features for time series classification problems. Traditional entropy measures, such as Shannon entropy, use probability distribution function. However, for the effective separation of time series, new entropy estimation methods are required to characterize the chaotic dynamic of the system. Our concept of Neural Network Entropy (NNetEn) is based on the classification of special datasets in relation to the entropy of the time series recorded in the reservoir of the neural network. NNetEn estimates the chaotic dynamics of time series in an original way and does not take into account probability distribution functions. We propose two new classification metrics: R2 Efficiency and Pearson Efficiency. The efficiency of NNetEn is verified on separation of two chaotic time series of sine mapping using dispersion analysis. For two close dynamic time series ($r$ = 1.1918 and $r$ = 1.2243), the F-ratio has reached the value of 124 and reflects high efficiency of the introduced method in classification problems. The electroencephalography signal classification for healthy persons and patients with Alzheimer disease illustrates the practical application of the NNetEn features. Our computations demonstrate the synergistic effect of increasing classification accuracy when applying traditional entropy measures and the NNetEn concept conjointly. An implementation of the algorithms in Python is presented.

## 1. Introduction

During the past 160 years, the concept of entropy has been applied to thermodynamical systems [1]. Over the years, the concept of entropy has been extended in various ways to solve scientific problems related to biomedical, healthcare, thermodynamics, physics, and others. Using horizon entropy, Jacobson and Parentani assessed the energy of black hole horizons [2]; Bejan studied entropy from the thermodynamic standpoint [3]. Bagnoli described the concept of entropy using the relationship between a thermal engine and a waterwheel [4]. His results showed that the concept of entropy is not restricted to thermal engines but can be applied to a wider variety of machines as well. A distribution entropy measure was used by Karmakar et al. for detecting short-term arrhythmias in heart rate [5].

There are two types of extended entropy measures: thermodynamic entropy and Shannon entropy. In thermodynamics, the entropy measure is related to the energy of a physical system. A Shannon entropy measure is an entropy measure that is used in information theory. Information/Shannon entropy measures are used to quantify the degree of freedom (DoF) or complexity of time series in practical applications. The technological advances in digitalization and their wide applications in practical problems in recent decades have made information entropy measures very popular.

The approximate entropy (ApEn) measure is a well-known entropy measure employed in biosignal analysis [6]. To quantify the self-similarity pattern in the given time series, ApEn considers the embedding dimension $m$, the time delay *tau* ($T$), and the tolerance value $r$ as the initial parameters. On the basis of initial parameters, a new trajectory matrix is constructed. Using the tolerance $r$, ApEn is computed by finding the self-similarity value between the rows of the trajectory matrix [6]. In short-length time series, ApEn is not accurate due to a strong dependence on data length [1]. Sample entropy (SampEn) and permutation entropy (PermEn) are introduced independently to overcome these difficulties [1]. The SampEn algorithm ignores self-matching patterns, which reduces computation time. It has been widely used in practical applications to analyze heart rate [7], recognize emotions [8], and analyze MRI signals [9]. The SampEn measure is inefficient when the time series are erratic or spiked [10].

In order to overcome these limitations, the Heaviside function, which is used in the SampEn algorithm, is substituted by a fuzzy membership function [1]. This new measure of entropy is called fuzzy entropy (FuzzyEn). According to experimental results, FuzzyEn performs better than SampEn [11,12]. In recent years, FuzzyEn has been used to detect epileptic seizures [13], recognize emotions [14], and investigate the dynamic of lung cancer images [15]. Chanwimalueang and Mandic proposed the CoSiEn as another improvement to SampEn [10]. They used a different (angular) metric to search for similar embedding vectors. Their results showed that CosiEn outperformed SampEn and FuzzyEn in the present 1/$f$ noise or short length dataset. In a study by Li et al. [16], the distribution entropy (DistEn) was introduced as an improvement over SampEn. According to Karmakar et al., DistEn is less sensitive than SampEn under initial parameters and gives additional information about time series [5]. Hence, DisEn can be regarded as an independent feature of SampEn.

Permutation entropy (PermEn), the other improvement of ApEn, combines entropy with symbolic dynamics [17]. In the PermEn algorithm, partitions are constructed by comparing neighboring time series values. PermEn is widely used in literature due to its robustness and computational efficiency [18]. Despite some advantages, PermEn is not stable regarding its input parameters. Therefore, the accuracy of PermEn depends on several input parameters. Its efficiency is affected by inconsistent results due to different initial parameters [19]. In order to overcome this difficulty, bubble entropy (BubbleEn) was introduced [19]. Similar to PermEn, BubbleEn uses a bubble sort algorithm to rank embedded vectors. However, BubbleEn is less sensitive to initial parameters than PermEn [19]. Increment entropy (IncEn) modifies PermEn to transform ranked time series into symbol sequences [20]. Consequently, IncEn is a suitable measure of entropy in the case of short datasets [20].

In addition to SampEn, PermEn, and their improved entropy measures, some entropy measures directly improve ShEn. Alter et al. extended ShEn for $A_{m \times n}$ matrices based on the singular value decomposition theorem (SVD) [21,22]. Singular value decomposition entropy (SVDEn) is defined as the ShEn of the normalized eigenvalues of the diagonal matrix (which appears in SVD). A new measure of entropy based on Poincare plots of time series was introduced by Yan et al. [23]. The Poincare plot is divided into $n \times n$ grids. In each grid, the probability of the number of each point is calculated. A measure of ShEn with respect to probability distributions is known as grid entropy (GridEn) [23]. Rohila and Sharma proposed phase entropy (PhaseEn), based on GridEn's concept [24]. In PhaseEn, the PhaseEn measure is computed using a second-order Poincare plot and ShEn [24]. In a recent study, Yang et al. introduced attention entropy (AttnEn) which considers only peak points. The results show that AttnEn is robust for short time series [25].

All of the mentioned entropy measures are extensions of the ShEn algorithm, which requires probability density functions. A measure of entropy called NNetEn [26] was introduced by Velichko and Heidari that uses feedforward neural networks with LogNNet models [27,28] and does not take into account probability distribution functions. In [29], the researchers showed that NNetEn is robust under noise conditions and outperforms traditional entropy measures.

Figure 1 shows the general concept of calculating NNetEn. A time series $X = (x_1…x_n)$, of length $N$, is written to the reservoir (stage 1). In the second stage, a dataset is selected to be used as the basis for calculating the classification metric. We feed the $Y$ vector from the dataset into the reservoir input (stage 3), and then pass it through normalization (stage 4). A time series element transforms the $Y$ vector in the reservoir (stage 5). Consequently, the output vector $Sh$ (stage 6) has a different dimension from the input vector $Y$. The output vector $Sh$ is normalized (stage 7) and fed into the output classifier (stage 8). In stage 9, the classifier is trained on the dataset, and in stage 10, it is tested. A linear transformation is performed on the classification metric in order to convert it into NNetEn entropy (Stage 11). In Section 2.3 of the methodology, the details of the steps are explained. A higher degree of irregularity in the time series $X$ leads to a more efficient process for transforming the input vector $Y$ into a different dimension, which in turn leads to a higher classification metric, and a higher output entropy. This principle differs from the common principles for calculating entropy based on a probability distribution.

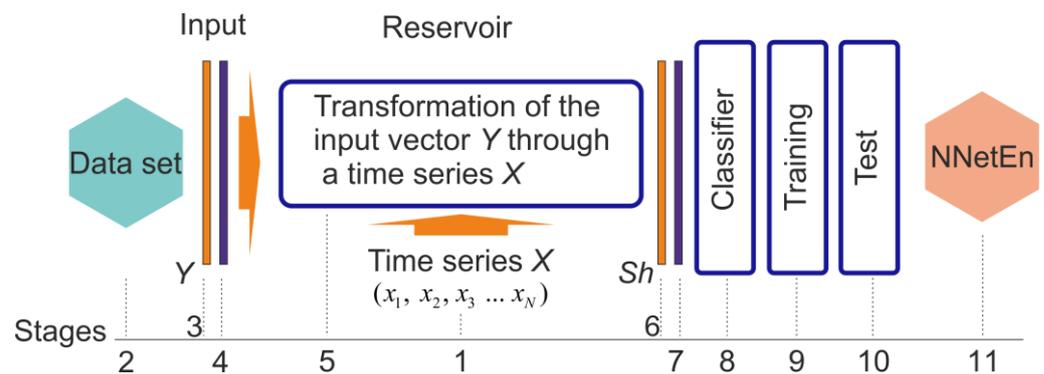

**Figure 1.** NNetEn calculation concept. The concept demonstrates the sequence of steps for computing NNetEn, including writing a time series to the reservoir, transforming the database in the reservoir through the time series, training and testing the output classifier, converting classification metric into NNetEn entropy.

The determination of data for entropy estimation (stage 2 in Figure 1) is an important aspect of the study. Initially, LogNNet was developed to compute the NNetEn based on MNIST handwritten dataset [30]. Here, we used 60,000 handwritten images for training and 10,000 handwritten images for testing the proposed entropy accuracy. Several studies have tested this method successfully [29,31–35]. Li et al. considered NNetEn as the characteristic of ship-radiated noise signals [31]. Heidari used NNetEn of electroencephalogram (EEG) signals for classifying motor imaginary of end users with disability [32]. He found that eight channels are enough for the classification, while other existed methods considered 30 channels. Velichko et al. introduced two-dimensional NNetEn for remote sensing imaginary and geophysical mapping [33]. NNetEn successfully performed analyzing the dynamic of chaotic spike oscillator based on S-switch [34] and measuring the complexity of response of traveling ionospheric disturbances under geomagnetic storms [35].

One of the disadvantages of the method is the high computation time associated with training and testing the network using the MNIST-10 database (stages 9, 10). To reduce computational overhead, we tested the method to reduce MNIST database size and used SARS-CoV-2-RBV1 dataset [36] as the input dataset in Stage 2.

Compared to the other entropy measures, the NNetEn has a unique feature to compute the dynamic behavior of the input data of either short or longer length and gives a meaningful information about the input data. In previous studies, NNetEn was computed using Delphi software, however, it is impractical to expect the researchers to be familiarized with this software tool. Hence, we aim to calculate the NNetEn measure using a widely used programming language Python. In this study, we have developed Python programming code to compute the NNetEn, tested the tool with different types of input data, and benchmarked the performance of the proposed method with the state-of-the-art methods in the literature.

The NNetEn's ability to differentiate signals was investigated on synthetic signals based on sinus mapping. In addition, we analyzed the EEG recordings obtained at AHEPA General Hospital of Thessaloniki's Department of Neurology [37]. Information on the functioning of the human brain can help to develop such tools as human-machine interface (HMI) and clinical diagnosis systems (CDS) [38]. Besides analyzing cognitive processes (emotional state, mental fatigue level, emotional stress, mental health), it is possible to detect the presence of a number of neurological diseases (epilepsy, Alzheimer's disease, attention deficit/hyperactivity disorder, bipolar disorder, etc.) [38]. In order to classify EEG signals, it is necessary to extract features from the original EEG signal. Methods to extract features include spectral analysis [39], wavelet analysis [40], entropy [41], statistical analysis [42], and others. Following the feature extraction, signals are classified using machine learning algorithms, such as a nearest neighbor method, a support vector machine, a decision tree, a random decision forest, neural networks, etc. [38,43]. In this paper, the robustness of NNetEn feature in analyzing EEG signal is investigated by considering AHEPA dataset. The AHEPA dataset includes electroencephalograms from 88 patients divided into three groups: control group (29 people), dementia (23), and Alzheimer's disease patients (36 people). Later, the NNetEn extracted from the above dataset has been processed and classified through a conventional machine learning algorithm (Support Vector Machine (SVM)). The performance of the NNetEn feature was analyzed with other conventional entropy measures. In addition to the existing accuracy metrics [44,45], we have developed two alternative classification accuracy metrics using $R^2$ and Pearson coefficients (R2 Efficiency and Pearson Efficiency) to assess the performance of the NNetEn.

The major contributions of the paper are:

- The concept of computing NNetEn is introduced.
- Investigating the effect of input dataset (Stage 2) in NNetEn value by considering the SARS-CoV-2-RBV1 dataset.
- Proposing the R2 Efficiency and Pearson Efficiency as new time series features related to NNetEn.
- Python package for NNetEn calculation is developed.
- The results of the separation of synthetic and real EEG signals using NNetEn are presented.
- The synergistic effect of using NNetEn entropy along with traditional entropy measures for classifying EEG signals is demonstrated.

The rest of the paper is organized as follows. In Section 2, the used datasets, proposed methods and new metrics are described. Section 3 is devoted to numerical results. The sine chaotic map and EEG signal of control group and patient with Alzheimer disease are analyzed. We demonstrate the proposed method, and the new metrics are robust for classifying different signals. Section 4 concludes the study and outlines directions for future research.

## 2. Materials and Methods

*2.1. Description of Datasets*

In this paper, we consider MNIST-10 [30] and SARS-CoV-2-RBV1 [36] as the input datasets for Stage 2 and the EEG dataset [37] as the input dataset for Stage 1 in the NNetEn algorithm. For simplicity, we call MNIST-10, SARS-CoV-2-RBV1 and the considered EEG dataset as Dataset 1, Dataset 2 and Dataset 3 respectively.

- Dataset 1**:** The MNIST dataset contains handwritten numbers from '0' to '9' with 60,000 training images and 10,000 testing images. Each image has a size of 28 × 28 = 784 pixels and is presented in grayscale coloring. This dataset is well balanced since the number of vectors of different classes is approximately the same. The distribution of elements in each class is given in Table 1.

**Table 1.** Number of images of different classes in MNIST-10 dataset.

| Classes | Number of Training Images | Number of Testing Images |
|---|---|---|
| 0 | 5923 | 980 |
| 1 | 6742 | 1135 |
| 2 | 5958 | 1032 |
| 3 | 6131 | 1010 |
| 4 | 5842 | 982 |
| 5 | 5421 | 892 |
| 6 | 5918 | 958 |
| 7 | 6265 | 1028 |
| 8 | 5851 | 974 |
| 9 | 5949 | 1009 |
| Total | 60,000 | 10,000 |

- Dataset 2: The SARS-CoV-2-RBV1 dataset contains information on 2648 COVID-19 positive outpatients and 2648 COVID-19 negative outpatients (control group), for 5296 patients, as described by Huyut, M.T. and Velichko in [36]. A data vector containing 51 routine blood parameters is included in this dataset. The dataset can be classified into two classes using binary methods such as COVID positive or normal control (NC). We used the entire dataset for both training and testing. As a result, the training and test sets coincided, and the resulting accuracy is equivalent to the accuracy on the training data.

Here, Dataset 1 and Dataset 2 are balanced so that simple metrics such as classification accuracy can be successfully applied to them.

In the study of NNetEn, we also measured the dependence of the signal separation efficiency on the database usage fraction $\mu$. For example, $\mu = 0.01$ means using 600 samples for training and 100 samples for testing for Dataset 1, and 53 samples for Dataset 2. Varying this parameter allows you to vary the entropy calculation time and its functionality.

- Dataset 3: This dataset contains records of EEG signals recorded at the AHEPA General Hospital of Thessaloniki's 2nd Department of Neurology [46]. This dataset consists of electroencephalograms of 88 patients divided into three groups: controls (29 people), Alzheimer's disease patients (36 people), and dementia patients (23 people). In order to record EEG, the authors of the dataset used a Nihon Kohden EEG 2100 device with 19 electrodes (channels) located on the head according to the 10–20 scheme: Fp1, Fp2, F7, F3, Fz, F4, F8, T3, C3, Cz, C4, T4, T5, P3, Pz, P4, T6, O1, O2. Each channel's signal was digitized at a sampling rate of 500 Hz. The duration of EEG recordings ranged from 5 min to 21.3 min. A resting EEG was recorded with the eyes closed.

## 2.2. Performance Metrics

Entropy is estimated using the classification metrics of neural networks: Classification Accuracy, R2 Efficiency and Pearson Efficiency.

- Classification Accuracy (*Acc*): Based on [26], we proposed to use the classification accuracy, which is as follows for multiclass classification:

$$\text{Metric: Accuracy} \quad Acc = \frac{\sum_{i=1}^{K} TP(C_{i,i})}{\sum_{i=1}^{K} \sum_{j=1}^{K} C_{i,j}}, \quad (1)$$

where $TP(C_{i,i})$ indicates the number of true positive classified elements in class $i$.

Here we use a multiclass confusion matrix with dimension $K*K$, where $K$ is the number of different class labels, a $C_{i,j}$ coefficients of confusion matrix [44].

There is only one disadvantage to using this metric, which is that it has a significant discreteness of value, which is especially noticeable when using a small dataset. If there are 100 vectors in the test dataset, then the precision value is limited to two decimal places (e.g., 0.55, 0.62, 0.99). In this paper, we present two more classification metrics that evaluate the efficiency of the classification algorithm by analogy with regression problems. These precision values have many significant digits after the decimal point. Figure 2 is presented for clarification. There is a data vector in the output layer of the classifier *Sout*, and a label vector in the training dataset *L*. A Pearson correlation coefficient (*p*) (Equation (2)) or a determination coefficient ($R^2$) (Equation (3)) is calculated for each pair.

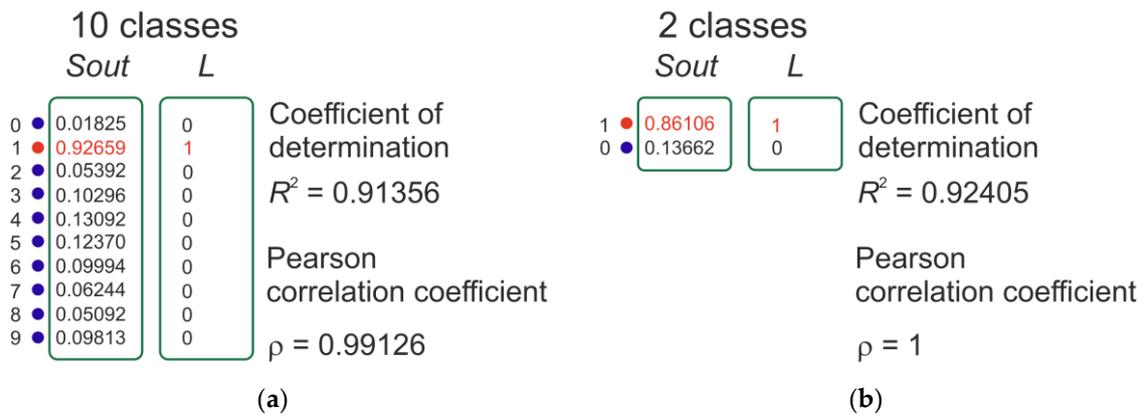

(a)  (b)

**Figure 2.** An example of calculating the determination coefficient $R^2$ and the Pearson correlation coefficient $\rho$ for a database with 10 classes of the MNIST dataset (**a**) and the SARS-CoV-2-RBV1 dataset with binary classification (**b**).

$$\text{Pearson correlation coefficient:} \quad \rho = \frac{\sum_{i=1}^{K}(L_i - \overline{L})(Sout_i - \overline{Sout})}{\sqrt{\sum_{i=1}^{K}(L_i - \overline{L})^2}\sqrt{\sum_{i=1}^{K}(Sout_i - \overline{Sout})^2}} \quad (2)$$

$$\text{Determination coefficient (R2):} \quad R^2 = 1 - \frac{\sum_{i=1}^{K}(L_i - Sout_i)^2}{\sum_{i=1}^{K}(L_i - \overline{L})^2} \quad (3)$$

Where $\overline{L}$ mean value of vector $L$, $\overline{Sout}$ mean value of vector *Sout*.

- R2 Efficiency (*R2E*):

The classification efficiency metric R2 Efficiency is proposed as an additional performance measure on this work. This is equal to the average $R_i^2$ over all vectors in the testing dataset with the number of vectors *M*.

$$\text{Metric: R2 Efficiency} \qquad R2E = \frac{\sum_{i=1}^{M} R_i^2}{M} \qquad (4)$$

The more closely the *Sout* vector repeats the label vector *L*, the closer R2 Efficiency is to 1.

- Pearson Efficiency (*PE*):

The Pearson Efficiency metric is proposed, which is equal to the average $\rho_i$ for all vectors in the testing dataset.

$$\text{Metric: Pearson Efficiency} \qquad PE = \frac{\sum_{i=1}^{M} \rho_i}{M} \qquad (5)$$

The more closely the *Sout* vector repeats the *L* label vector, the closer the Pearson Efficiency is to 1.

As a result, NNetEn is equivalent to the value of one of the three types of metrics.

$$\text{NNetEn} = \begin{cases} \text{Accuracy} \\ \text{R2 Efficiency} \\ \text{Pearson Efficiency} \end{cases} \qquad (6)$$

The metric type is an input parameter to the entropy function (see Section 2.4).

*2.3. NNetEn Calculation*

As shown in Figure 3, the main steps involved in calculating NNetEn are described in detail. Basically, it is the same structure as LogNNet neural networks we presented in [27]. An important feature of the reservoir is the presence of a matrix transformation, in which the matrix coefficients are filled with the time series *X*.

The main stages of NNetEn calculation used in this study are described in detail below.

Stage 1: The first stage involves writing the time series $X = (x_1…x_n)$, length *N*, into the reservoir. The maximum length of the series *Nmax* is determined by the number of elements in the reservoir matrix $Nmax = Y\_max \times P\_max$, where $Y\_max$ refers to the dimension of the *Y* vector of the dataset and $P\_max$ refers to the dimension of the vector *Sh* (the first layer) of the output classifier; we used $P\_max = 25$ in our work [29]. For Dataset 1, $N\_max = (784 + 1) \times 25 = 19{,}625$ and for Dataset 2, $N\_max = (51 + 1) \times 25 = 1300$.

In our earlier work, there are six main methods available for filling the reservoir are explored, detailed in [29]. The following conventions apply to methods M1…M6: M1—Row-wise filling with duplication; M2—Row-wise filling with an additional zero element; M3—Row-wise filling with time series stretching; M4—Column-wise filling with duplication; M5—Column-wise filling with an additional zero element (see Figure 3); M6—Column-wise filling with time series stretching.

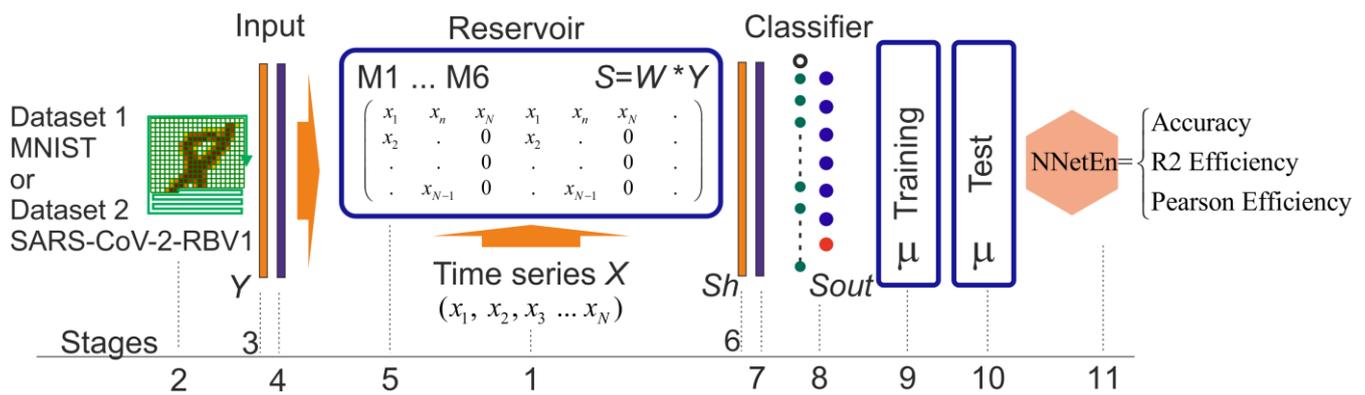

**Figure 3.** Main steps of NNetEn calculation.

Stage 2: Datasets 1 and 2 are selected based on which the classification metric will be calculated.

Stage 3: The *Y* vector is formed from the dataset, including a zero offset $Y[0] = 1$. For Dataset 1 vector dimension is $Y\_max = 784 + 1$, for Dataset 2 $Y\_max = 51 + 1$.

Stage 4: The *Y* vector goes through the normalization stage in stage 4. For Dataset 1, this is the division of all elements (except $Y[0]$) by 255. Based on the maximum and minimum values in the original database, *Y* is normalized for Dataset 2.

Stage 5: In stage 5, the vector *Y* has transformed into a vector *Sh* by multiplying it with the reservoir matrix and the input vector $Sh = W \times Y$ (see Algorithm 1).

Stage 6: In stage 6, Vector *Sh* enters the input layer of the classifier with a dimension of $P\_max = 25$.

Stage 7: The vector *Sh* goes through the normalization stage, as described in Algorithm 1. Here is a comparison of the executable code in Delphi (Algorithm 1a) and Python (Algorithm 1b), which uses vectorized calculations. In this array, *Sh_min*, *Sh_max*, and *Sh_mean* [27] are the minimum, maximum, and average values of the vector *Sh* calculated across the entire database.

Stage 8: A single-layer output classifier with ten neurons for Dataset 1 and two neurons for Dataset 2 (see Figure 2). An activation function based on logistic regression was used.

Stage 9–Stage 10: The training and testing phases of the neural network are shown in stages 9–10. The training was carried out using the backpropagation method, with a variable number of epochs (*Ep*). The value of *Ep* is a parameter of the entropy function.

Stage 11: In accordance with equation 6, the classification metric is transformed linearly into the NNetEn entropy based on one of three options.

**Algorithm 1.** Implementation of the calculation of the values of the vector *Sh* and its normalization in Delphi—(**a**), Python—(**b**).

| (a) | (b) |
|---|---|
| ```for j := 1 to P_max do``` | ```Sh = np.sum(np.multiply(Y, W), axis=1)``` |
| ```begin``` | ```Sh = np.divide(Sh - Sh_min,``` |
| ```    Sh[j] := 0;``` | ```                  Sh_max - Sh_min)``` |
| ```    for i := 0 to Y_max do``` | ```     - 0.5 - Sh_mean``` |
| ```        Sh[j] := Sh[j] + Y[i] * W1[i,j];``` | |
| ```    if (Sh_max[j] - Sh_min[j]) <> 0 then``` | |
| ```        Sh[j] := ((Sh[j] - Sh_min[j]) /``` | |
| ```                  (Sh_max[j] - Sh_min[j])``` | |
| ```                  -0.5) - Sh_mean[j];``` | |
| ```end;``` | |

*2.4. Entropy Settings Options in Python*

The calculation for SampEn, CoSiEn, FuzzyEn, PhaseEn, DistEn, BubbleEn, GridEn, IncEn and AttnEn entropy measures was implemented using the EntropyHub [47] software package. SVDEn and PermEn were calculated using the Antropy [48] software package.

The following parameters were used to calculate the entropies:

- SVDEn ($m = 2$, *delay* = 1);
- PermEn ($m = 4$, *delay* = 2);
- SampEn ($m = 2$, $r = 0.2 \cdot d$, $\tau = 1$), where $d$ is standard deviation of the time series;
- CoSiEn ($m = 3$, $r = 0.1$, $\tau = 1$);
- FuzzyEn ($m = 1$, $r = 0.2 \cdot d$, $r_2 = 3$, $\tau = 1$), where $d$ is standard deviation of the time series and $r_2$ is fuzzy membership function's exponent;
- PhaseEn ($K = 3$, $\tau = 2$);
- DistEn ($m = 3$, *bins* = 100, $\tau = 1$);
- BubbleEn ($m = 6$, $\tau = 1$);
- GridEn ($m = 10$, $\tau = 1$);
- IncEn ($m = 4$, $q = 6$, $\tau = 1$);
- AttnEn has no parameters;

The following specifiers are included in parentheses for NNetEn settings:

NNetEn (Database (D1—Dataset 1 or D2—Dataset 2), database usage fraction $\mu$, reservoir filling method (M1–M6), number of epochs ($Ep$), classification metric type (Equation (6))). For example, NNetEn(D1, 1, M1, Ep5, R2E) uses Dataset 1, $\mu = 1$—full base, reservoir filling method M1, number of epochs $Ep = 5$, classification metric R2 Efficiency).

There are 72 gradations of settings for Dataset 1 and Dataset 2.

*Nset* settings are numbered according to the following formula:

$$Nset = (m_1 - 1) \cdot 24 + (m_2 - 1) \cdot 4 + m_3$$

$$m_1 = \begin{cases} 1, & \text{if Metric = R2 Efficiency} \\ 2, & \text{if Metric = Pearson Efficiency} \\ 3, & \text{if Metric = Accuracy} \end{cases}$$

$m_2 = 1...6$, number of matrix method filing method M1...M6 \hfill (7)

$$m_3 = \begin{cases} 1, & \text{if } Ep = 1 \\ 2, & \text{if } Ep = 5 \\ 3, & \text{if } Ep = 20 \\ 4, & \text{if } Ep = 100 \end{cases}$$

Each group of settings *Nset* = 1, …, 24 is analyzed NNetEn using the R2E metric, *Nset* = 25, …, 48 is analyzed NNetEn using the Pearson Efficiency metric, and *Nset* = 49, …, 72 is analyzed NNetEn using the Accuracy metric. The three groups are further divided into six subgroups that fill the reservoir in different ways according to the number of epochs *Ep*.

*2.5. Generation of Synthetic Time Series*

To generate synthetic time series, we used the discrete chaotic sine map [26]:

$$x_{n+1} = r \cdot \sin(\pi \cdot x_n), \quad 0.7 \leq r \leq 2, \, x_{-999} = 0.1, \tag{8}$$

The first 1000 elements are ignored due to the transient period. If $n > 0$, then the NNetEn measure is calculated for $x_n$. In this series, $N = 300$ elements were included. To generate a class corresponding to one value of $r$, 100-time series were generated. Elements in each series were calculated sequentially equation 8, ($x_1$, …, $x_{300}$), ($x_{301}$, …, $x_{600}$), etc. Figure 4a shows an example of a bifurcation diagram for a sine map.

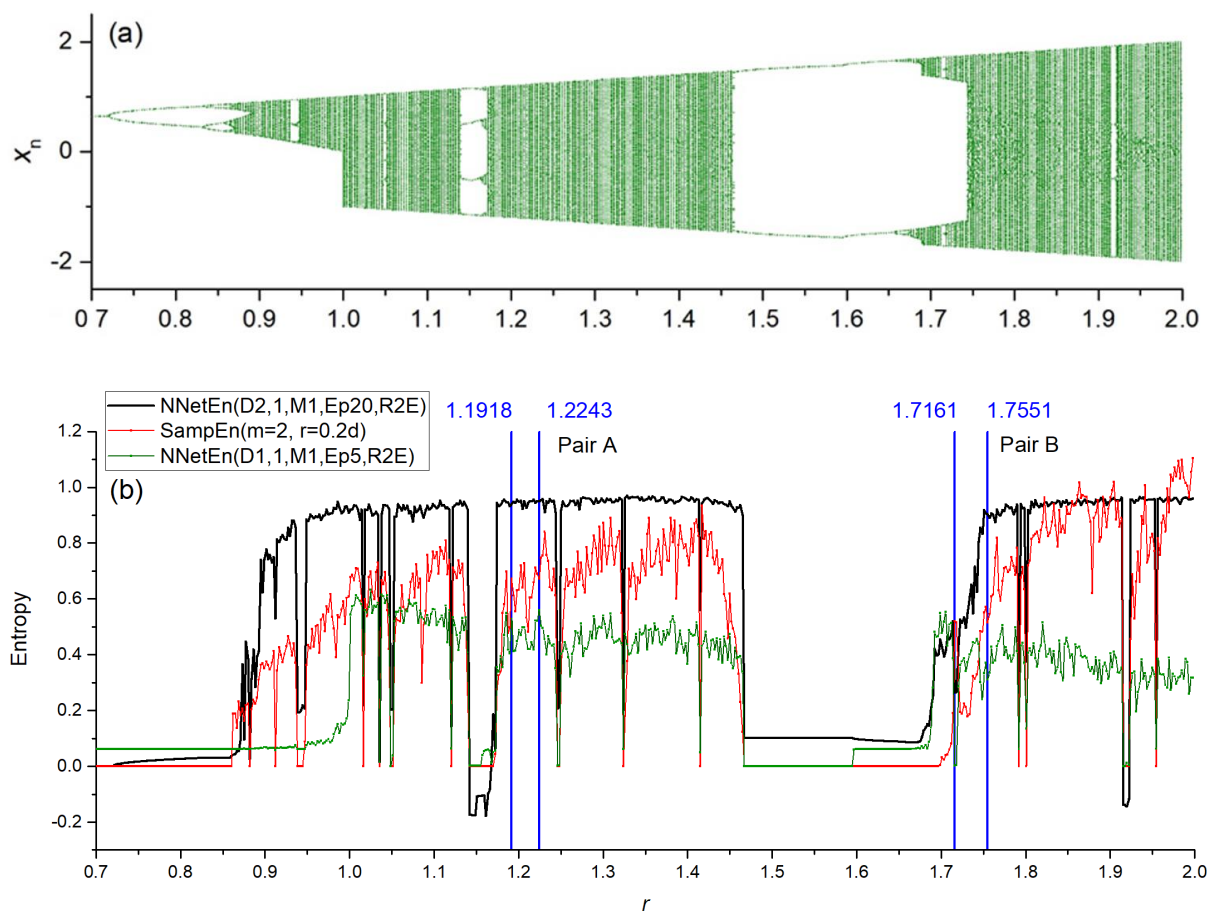

**Figure 4.** (**a**) Bifurcation diagrams for sine map (Equation (6)); (**b**) the dependence of entropy on the parameter *r* for NNetEn and SampEn. Comparison of figures (**a**) and (**b**), shows that the ranges *r* that demonstrate regular oscillations have low entropy values, while ranges with chaotic oscillations have high entropy values.

Figure 4b shows the NNetEn (black color for D2 and green color for D1) dependencies for the two parameters, along with an example of SampEn (red color). While sections with regular oscillations show decreased entropy values, those with chaotic oscillations have increased entropy values. It is evident that the form of dependencies differs greatly.

We computed the NNetEn measure for pair A (*r* = 1.1918 and 1.2243) and pair B (*r* = 1.7161 and 1.7551). Their mutual arrangement is shown in Figure 4b. Figure 5 shows examples of signals for pair A, and Figure 6 shows examples for pair B. As can be seen, pair A is a chaotic oscillation, and it is difficult to distinguish between the signals visually in terms of the amount of randomness and the differences in dynamics between them. In pair B, the differences in signals are more pronounced. Compared to the periodic signal with *r* = 1.7161, the chaotic signal with *r* = 1.7551 changes its amplitude chaotically, mostly in the same region. For each class, the mean entropy value NNetEn$_{av}$ was determined for these settings, as well as the standard deviation *S*.

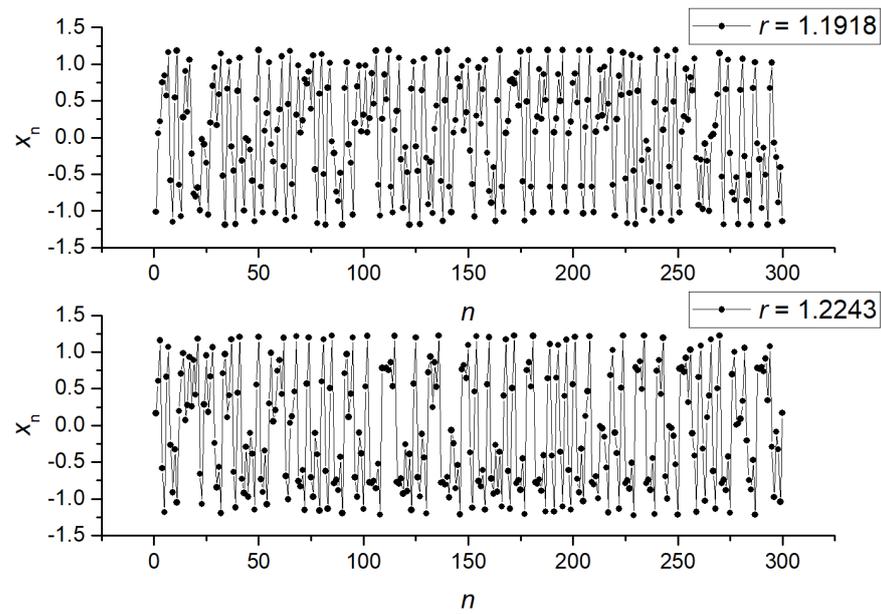

**Figure 5.** Examples of signals ($x_1$, ..., $x_{300}$) for pair B ($r = 1.1918$ and $r = 1.2243$).

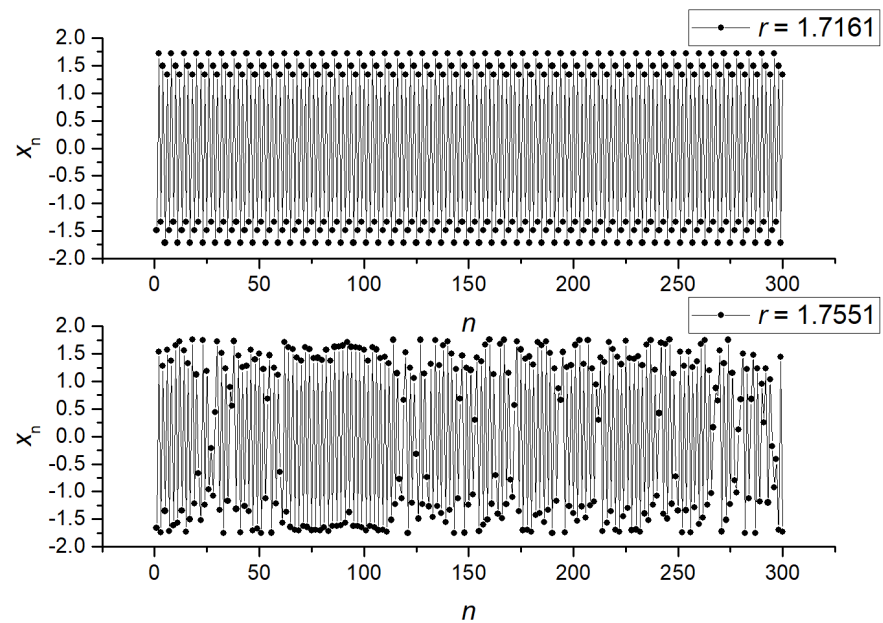

**Figure 6.** Examples of signals ($x_1$, ..., $x_{300}$) for pair B ($r = 1.7161$ and $r = 1.7551$).

*2.6. Signal Separation Metrics*

2.6.1. Statistical Analysis of Class Separation

In order to assess the degree of class separation, analysis of variance (ANOVA) was performed, with a significance level of $p < 0.05$. ANOVA allows us to compare the variances of the means of different groups [45]. The F-ratio measures the difference between variance within and variance between groups. The entropies of each signal within the class were calculated, and then the F-ratio between the classes of pair A and pair B of synthetic signals, as well as between the control group and the group of patients with Alzheimer's disease, was calculated (Section 3.4). As a result, the larger the F-ratio, the stronger the separation between two classes of signals.

2.6.2. Calculation of the Accuracy of Signal Classification by Entropy Features

A Support Vector Machine (SVM) was used to classify the signal using one and two entropy features.

Two steps were involved in evaluating the classification metric.

- In the first stage, hyperparameters were selected using Repeated K-Fold cross-validation (RKF). In order to accomplish this, the original dataset was divided into $K = 5$ folds in $N = 5$ ways. The K-folds of each of the $N$ variants of partitions were filled differently with samples. In addition, the distribution of classes in each K-fold approximated the distribution in the original dataset. Next, the classifier hyperparameters were selected at which the average accuracy of the classifier on the validation set was maximized. As a result of using a large number of training and validation sets, repeated K-Fold cross-validation allows minimize the overfitting of the model.
- As a second step, we used the hyperparameter values obtained in the first stage and performed RKF cross-validation ($K = 5$) in a similar manner to the first stage, but using different $N = 10$ partitions. As a result, the $A_{RKF}$ is calculated by averaging $N$ partitioned scores. We used $A_{RKF}$ as a metric to assess signal separation.

2.6.3. Synergy Effect Metric

As a result of combining entropy features during signal separation, $A_{RKF}$ classification becomes more accurate than if each feature were used individually. Using the formula below, the synergistic effect coefficient $K_{syn}$ was estimated.

$$Ksyn = \frac{1 - MAX(A_{RKF}[\text{Entropy1}], A_{RKF}[\text{Entropy2}])}{1.001 - A_{RKF}[\text{Entropy1}, \text{Entropy2}]} \quad (9)$$

where, $A_{RKF}$[Entropy1] is the classification accuracy using one entropy feature Entropy1, and $A_{RKF}$[Entropy1, Entropy2] is the classification accuracy using two entropy features. *MAX* is the maximum value selection function.

*2.7. Python Package for NNetEn Calculation*

2.7.1. General Requirements

The following libraries were used to implement the NNetEn algorithm in Python:

- NumPy
- Numba

NumPy contains all the necessary mathematical operations for matrices and arrays. This library offers optimization of operations and vectorization of calculations. The Numba compiler library converts Python functions into machine code at runtime using the industry-standard LLVM (Low-Level Virtual Machine) library [49]. Python version 3.7 or later is required to run the algorithm. There are four main blocks in the Python algorithm for calculating NNetEn, including those described in Section 2.3:

Stage 1 (Block 1): Reading time series from the input data files.

Stage 2–Stage 4 (Block 2): An instance of the NNetEn_entropy class is created, containing normalized, training, and test sets for the LogNNet neural network. The data at these stages are prepared for further analysis, as Datasets 1 and Datasets 2 have different formats. The MNIST dataset consists of four binary files: a training and test set of images, and a training and test set of labels. Training and test data are contained in one file in the SARS-CoV-2-RBV1 dataset. In addition, the $\mu$ parameter (database usage fraction) is passed here.

Stages 5–10 (Block 3): Calculation of classification accuracy by neural networks LogNNet. Several parameters need to be passed to the NNetEn_calculation function, including the reservoir formation method (M1, …, M6), the number of neural networks training epochs, and the metric calculation algorithm.

Stage 11 (Block 4): NNetEn entropy calculation with parameters written to log file. The log file contains the following data: the timestamp when the recording was made, the NNetEn value, the number of epochs, the size of the reservoir matrix $W$, $\mu$, and the length of the time series. The format makes it easier to analyze the data in the future.

2.7.2. Function Syntax

The program installation is done from PyPi repository using the following command (Listing 1).

**Listing 1.** Command to installation a NNetEn package.

```
1   pip install NNetEn
```

An instance of the NNetEn_entropy class is created by two commands (Listing 2).

**Listing 2**. Commands to create a NNetEn_entropy model.

```
1   from NNetEn import NNetEn_entropy

2   NNetEn = NNetEn_entropy(database = 'D1', mu = 1)
```

Arguments:
- database— (default = D1) Select dataset, D1—MNIST, D2—SARS-CoV-2-RBV1
- mu— (default = 1) usage fraction of the selected dataset $\mu$ (0.01, …, 1).

Output: The LogNNet neural network model is operated using normalized training and test sets contained in the NNetEn entropy class.

To call the calculation function, one command is used (Listing 3).

**Listing 3**. NNetEn calculation function with arguments.

```
1   value = NNetEn.calculation(time_series, epoch = 20, method = 3, metric = 'Acc', log = False)
```

Arguments:
- time_series—input data with a time series in numpy array format.
- epoch— (default epoch = 20). The number of training epochs for the LogNNet neural network, with a number greater than 0.
- method— (default method = 3). One of 6 methods for forming a reservoir matrix from the time series M1, ..., M6.
- metric —(default metric = 'Acc'). See Section 2.2 for neural network testing metrics. Options: metric = 'Acc', metric = 'R2E', metric = 'PE' (see Equation (6)).
- log— (default = False) Parameter for logging the main data used in the calculation. Recording is done in log.txt file

Output: NNetEn entropy value.

The source code of thePython package is stored on the site (https://github.com/izotov93/NNetEn (accessed on 26 April 2023)), and an example of the calculation is presented in Supplementary Materials.

## 3. Numerical Results and Discussion

### 3.1. Separation of Synthetic Signals

In Figure 7, statistics are calculated for two signals by choosing $r = 1.11918$ and $1.2243$ in Equation (8) which is called pair A. We present the dependence of the mean entropy value NNetEn$_{av}$ (with standard deviation $S$) based on $Nset$ setting numbers for Dataset 1 (Figure 7a left axis) and Dataset 2 (Figure 7b left axis). Furthermore, F-ratios are shown by blue diagrams based on the right vertical axis. Results for pair B ($r = 1.7161$ and $r = 1.7551$) are shown in Figure 8.

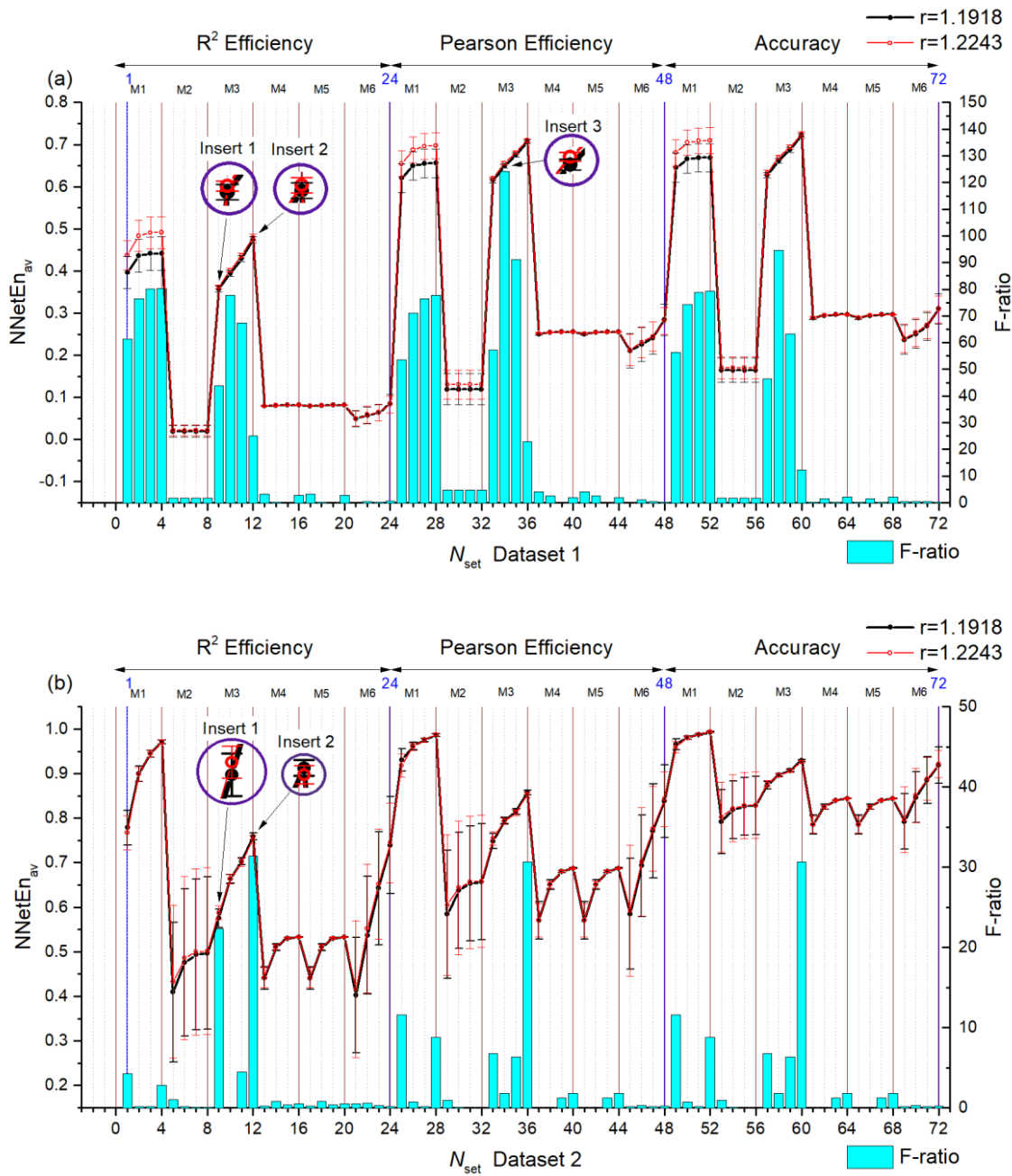

**Figure 7.** Dependences of the average entropy value NNetEn$_{av}$ (indicating the standard deviation $S$) and F-ratio depending on the $Nset$ setting number for Dataset 1 (**a**) and Dataset 2 (**b**). Pair A signals. The insets in the figure show the entropy ratio for the two settings.

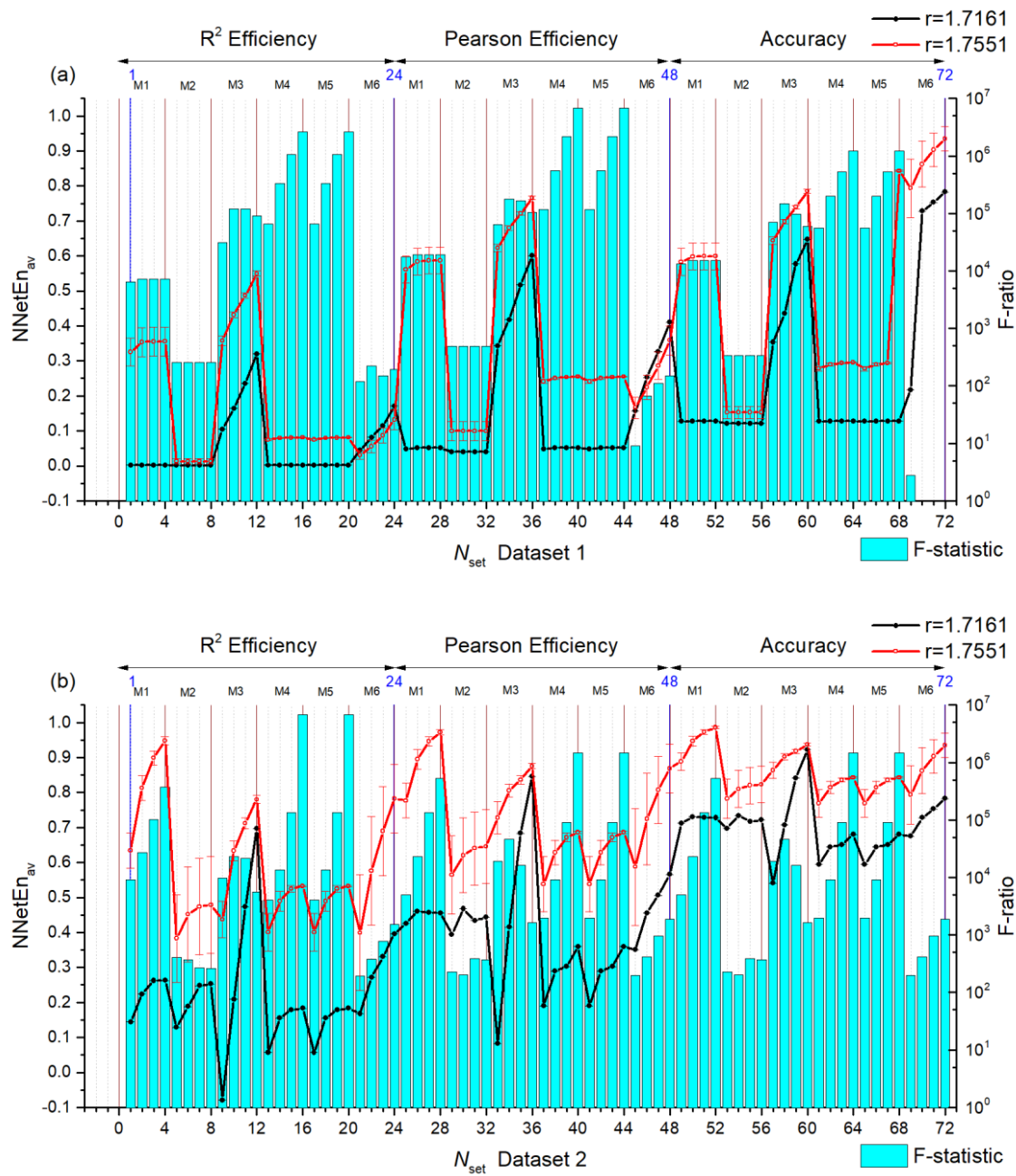

**Figure 8.** Dependences of the average entropy value NNetEn$_{av}$ (indicating the standard deviation *S*) and F-ratio depending on the *Nset* setting number for Dataset 1 (**a**) and Dataset 2 (**b**). Pair B signals.

There are several observations that can be made from the above results.

In general, the F-ratio value varies with the number of settings. For pair A, it reaches the maximum values of ~124 (Dataset 1, *Nset* = 34) and ~30 (Dataset 2, *Nset* = 12, 36, 60). Therefore, Dataset 1 shows a large separation capacity for given pairs of signals. In Dataset 1 higher F-ratio values correspond to methods of filling reservoirs M1 and M3. In Dataset 2 higher F-ratio values correspond to methods of filling reservoirs M1, M2, M3 and M5.

The higher the F-ratio, the more the NNetEn$_{av}$ values differ between the signals in the pair. This is clearly seen in insert 1 of pair A (*Nset* = 9) insert 2 (*Nset* = 12), and insert 3 (*Nset* = 34) when zooming in (see Figure 7a). As a result, NNetEn$_{av}$ shows the biggest difference, with a low *S* value being associated with *Nset* = 34. For pair B, this pattern is clearly visible, for example, with the (*Nset* = 28) and (*Nset* = 29) (Figure 8a).

As a result of the large difference in NNetEn$_{av}$ for Pair B, the F-ratios are extremely large at ~10$^6$.

For Dataset 1 and Dataset 2, there is a dependency that increases in $Ep$ lead to increases in NNetEn$_{av}$. An increase in the number of epochs leads to an increase in network training and database recognition efficiency, which is measured by NNetEn (Equation (6)). In addition, $S$ decreases with increasing epochs, i.e., entropy for a given signal reaches stationary values.

Several settings have been found that cause one entropy for one signal to exceed the entropy of another signal, while other settings cause the ratio to change. Therefore, Figure 7b for Dataset 2 shows setting where NNetEn$_{av}$ ($r$ = 1.2243) is greater than NNetEn$_{av}$ ($r$ = 1.1918) (for example, $Nset$ = 9 see insert 1), and setting where NNetEn$_{av}$ ($r$ = 1.2243) is less than NNetEn$_{av}$ ($r$ = 1.1918) (for example, $Nset$ = 12 see insert 2). Similarly, Figure 8a for Dataset 1 shows a change in ratios. A change in the ratios can be beneficial when applying combinations of entropies (see Section 3.2).

Entropy metrics (Accuracy, Pearson Efficiency, and R2 Efficiency) generally performed similarly, but F-ratios varied depending on the pair of signals. The Pearson Efficiency metric was in the lead for pair A and base D1 (Figure 7a). According to Figure 8a, R2 Efficiency led for pair B and base D2.

*3.2. Entropy Combinations*

Our previous section showed that NNetEn by itself can be a strong feature for splitting signals into pairs. However, using combinations of entropies should produce the greatest effect. In the following paragraphs, we discuss the results for two features, whose simplest combination is the difference. In order to separate the signals, a pair of A signals were used as a more complex pair.

3.2.1. Entropy Difference NNetEn as a Feature for Signal Separation

We can consider the relationship between the F-ratio and the difference in entropies. In Figure 9, the F-ratio distributions for three variants of difference are shown: NnetEn(D1)-NNetEn(D1) (Figure 9a), NnetEn(D2)-NNetEn(D2) (Figure 9b), NNetEn(D2)-NNetEn(D1) (Figure 9c). For the same time series of signals, the entropy difference was computed with different settings.

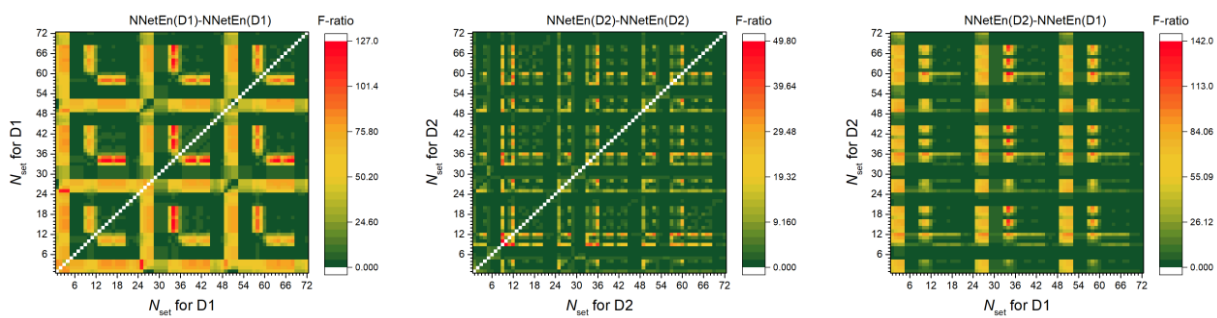

**Figure 9.** F-ratio distribution for three variants of entropy differences: NnetEn(D1)-NNetEn(D1) (**a**), NnetEn(D2)-NNetEn(D2) (**b**), NNetEn(D2)-NNetEn(D1) (**c**). The figure demonstrates the relationship between the F-ratio and the difference in entropies. The entropy difference under certain settings can have high efficiency (F-ratio) when used as a feature.

In Figure 9a, the maximum F-ratio (~127) is only 3 points higher when using only Dataset 1, whereas when using one feature. In Figure 9b, F-ratio reached a maximum of 50 using only Dataset 2, which is 20 points higher than when using one feature. As a result of combining both datasets (Figure 9c), the F-ratio reached 140, which is 13 points higher than all previous options combined. A feature difference is therefore even more powerful than an individual feature. It can be explained by a change in the relation between entropies. For example, for Pair A and Dataset 2 (Figure 7b) with $N_{set}$ = 9 NNetEn$_{av}$ ($r$ = 1.1918) < NNetEn$_{av}$ ($r$ = 1.2243), and for $N_{set}$ = 12 NNetEn$_{av}$ ($r$ = 1.1918) > NNetEn$_{av}$ ($r$ = 1.2243), therefore, the entropy difference at these settings changes more strongly from class to class. As a result, we see the maximum F-ratio for the point (9,12) in Figure 9b.

3.2.2. NNetEn as a Paired Feature in Signal Classification

The problem of separating signals is addressed through the definition of classification accuracy $A_{RKF}$ described in Section 2.6.3. Figure 10a shows the $A_{RKF}$ dependencies using the NnetEn(D1) and NnetEn(D2) features separately. NnetEn(D1) has a higher classification accuracy than NnetEn(D2), reaching the maximum of $A_{RKF}$ ~0.837 at $N_{set}$ = 34. It is consistent with Figure 7 where the F-ratio also reached its maximum at $N_{set}$ = 34, with Dataset 1 showing higher values than Dataset 2.

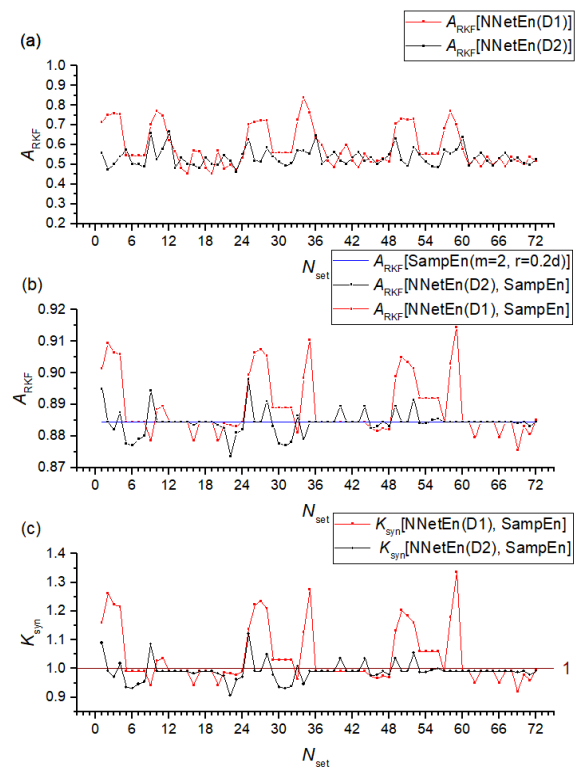

**Figure 10.** Classification accuracy $A_{RKF}$ dependences using NNetEn(D1) and NNetEn(D2) features separately (**a**), their combination with SampEn (**b**) and dependence of the synergistic effect coefficient (**c**). The figure shows the efficiency of using NNetEn as a feature pair to SampEn.

We used SampEn($m$ = 2, $r$ = 0.2$d$) together with NnetEn(D1) and NnetEn(D2). According to Figure 10b, a single SampEn gives recognition accuracy $A_{RKF}$ = 0.8845. By combining entropies, $A_{RKF}$ can either increase, reaching values of ~0.9145 or decrease to ~0.8735. The combination of SampEn and NNetEn entropies can significantly increase classification accuracy. Figure 10c illustrates a quantitative assessment of the synergistic effect. At $N_{set}$ = 2, 27, 35, 50, 59, the D1 base achieves the highest Ksyn. The settings numbers correspond to M1 and M3 reservoir filling methods. A two-dimensional diagram is shown in Figure 11 for the combination [NNetEn(D1, $N_{set}$ = 2), SampEn]. As can be seen, there is a selective extrusion of points along the ordinate axis as indicated by the sign NNetEn(D1, $N_{set}$ = 2). In this way, it is possible to more clearly separate the classes; in the figure, a blue straight line indicates a conditional separation between the classes. A slanted blue line separates classes better than a vertical or horizontal line, and it proves the effectiveness of using NNetEn and SampEn in pairs.

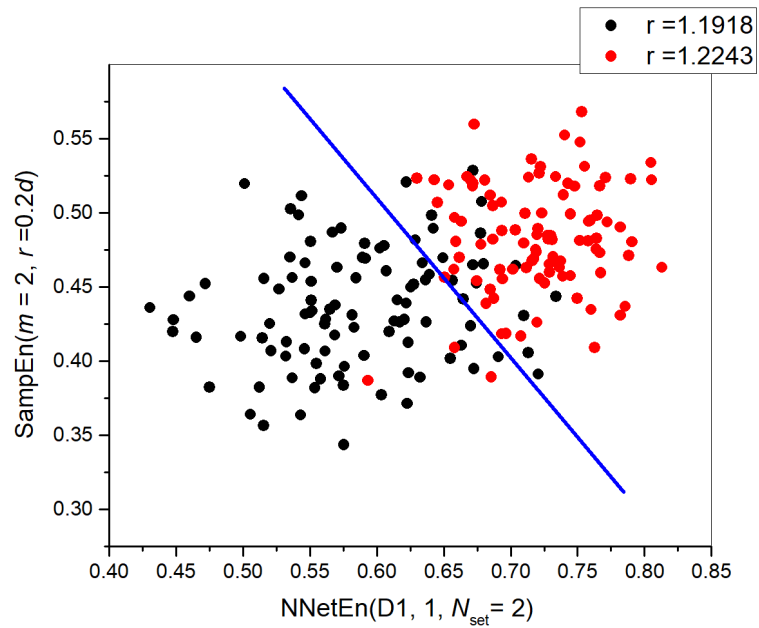

**Figure 11.** Feature combination diagram [NNetEn(D1, $N_{set}$ = 2), SampEn($m$ = 2, $r$ = 0.2d)]. The figure demonstrates the effectiveness of NNetEn and SampEn pairing, as a slanted blue line separates classes better than a vertical or horizontal one.

*3.3. Dependence of F-Ratio and Algorithm Speed on Dataset Size*

To vary the size of the dataset, the parameter $\mu$ (0.01, ..., 1) was introduced. It determines the fraction of the database usage. Figure 12 shows the dependences of F-ratio and calculation time of one time series on $\mu$.

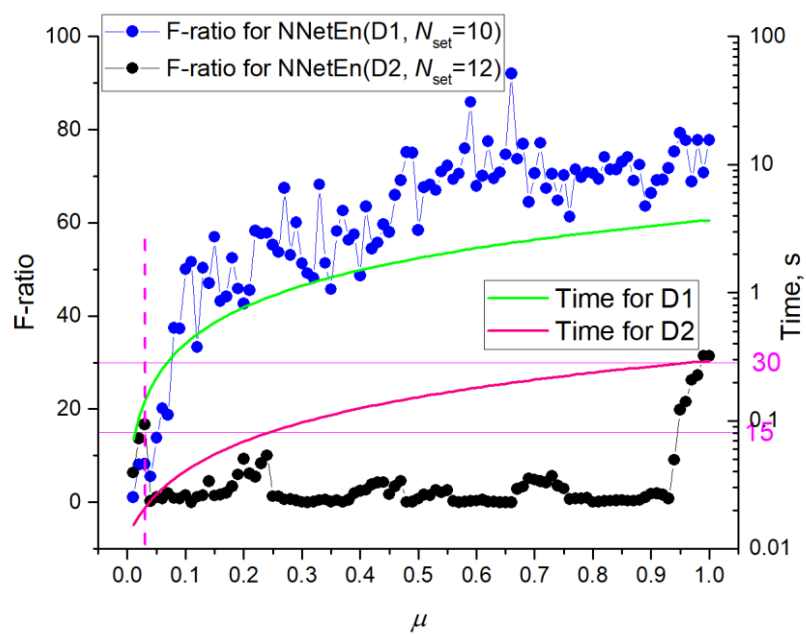

**Figure 12.** Dependences of F-ratio and calculation time of one time series on $\mu$. The figure shows that decreasing $\mu$ can significantly reduce entropy calculation time. For some settings, time is reduced without significant loss in classifying accuracy (F-ratio).

NNetEn with Dataset 1 shows a gradual decrease in F-ratio for $\mu > 0.2$ and an intense decrease for $\mu < 0.2$. Dataset 2 already at $\mu < 0.95$ has low F-ratio values, with the presence of local maxima. As a result of different database sizes, the algorithm for Dataset 2 takes approximately an order of magnitude less time to execute than Dataset 1. With a decrease in $\mu$, calculation time decreases. Based on the F-ratio of 30 and the average calculation time of 0.2 s, datasets have approximately the same efficiency regarding time costs. If we compare the time costs at the F-ratio ~15 level, then using Dataset 2 speeds up calculations per order, due to the presence of local maxima at a low value of $\mu$~0.03, where the calculation time for NNetEn(D1) is ~0.2 s and for NNetEn(D2) ~0.02 s.

In Figure 13, the dependences between $NNetEn_{av}(r1)$, a difference ($NNetEn_{av}(r2) - NNetEn_{av}(r1)$), and the standard deviation $S(r1)$ are plotted to reveal the reason for the sharp change in F-ratio. Here, $r1$ is 1.1918 and $r2$ is 1.2243. It can be seen that for Dataset 1 at $\mu > 0.2$, the standard deviation is approximately at the same level, and the difference in entropy slowly decreases. At $\mu < 0.2$, there is a sharp increase in $S(r1)$ and a sharp decrease in the difference of entropies. For Dataset 2, a sharp decrease in F-ratio at $\mu = 0.95$ is associated with a sharp decrease in the value of the entropy difference, which changes its sign for the first time at $\mu = 0.8$. At $\mu < 0.2$ for Dataset 2, there is a sharp increase in the difference between entropy and standard deviation.

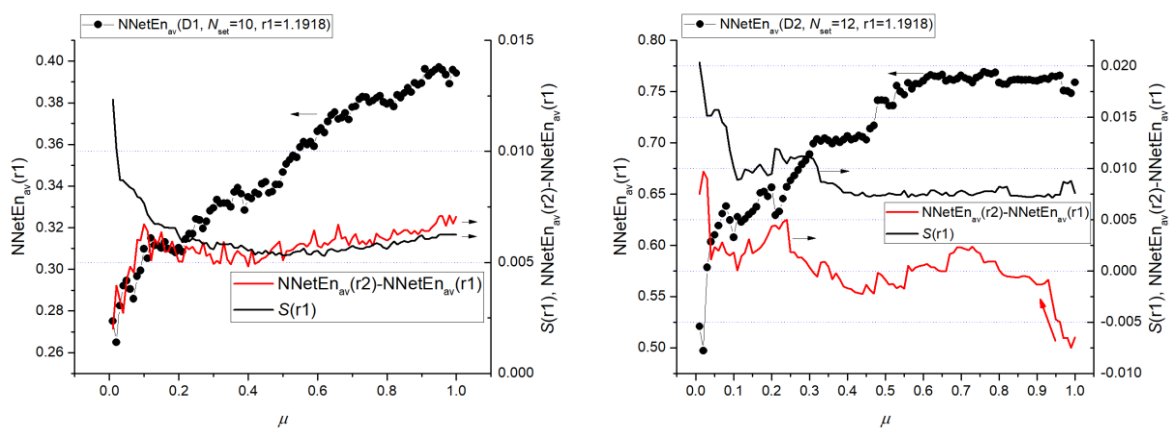

**Figure 13.** Dependences of entropy values $NNetEn_{av}(r1)$, their difference ($NNetEn_{av}(r2) - NNetEn_{av}(r1)$) and standard deviation $S(r1)$ for Dataset 1 (**a**) and Dataset 2 (**b**). Figure reflects the reason for the sharp change in F-ratio (in Figure 12).

*3.4. EEG Signal Separation*

3.4.1. Selection of the Most Informative Component of the EEG Signal

As described in Section 2.1, this work utilizes the Dataset 3 to study EEG signal separation. Due to the complexity of the problem, it was decided to use only the records of patients with Alzheimer's disease (36 people) and those in the control group (29 healthy people). Because the number of patients in these records is small, six non-overlapping segments were selected from each record, each of which represented an independent observation. All segments lasted 10 s and had the same duration. As the original dataset contained time stamps indicating unwanted patient activity (muscle movements, swallowing, blinking), the segments were selected to avoid overlap with the indicated activity periods.

The EEG signal's entropy may be a sign of poor separation ability since it has a wide spectral range (0–250 Hz), and the main information about brain activity lies in a relatively narrow set of frequencies [50]. The wavelet transform can be used to decompose the EEG signal into separate frequency components and increase the information content. A 5th-order digital Butterworth filter was used to pre-filter the signal (lower frequency 0.5 Hz, upper frequency 32 Hz) [51]. In this case, the discrete wavelet transform (DWT) of the signal was carried out using the *db*4 wavelet and performed six level decomposition.

A total of 14 options were considered to select the most informative component of the signal:

1: The original unfiltered $X_{EEG}(t)$ signal.

2: $XF_{EEG}(t)$ filtered signal.

3–8: Approximation coefficients (A1–A6) of the wavelet transform of the filtered signal.

9–14: Detail coefficients (D1–D6) of the wavelet transform of the filtered signal.

Then, 14 datasets were created, each containing 390 records (65 patients, 6 segments) and 19 features—the SVDEn ($m = 2$, *delay* = 1) value for 19 channels, followed by a class label of 0 indicating a control group patient and 1 indicating a patient with Alzheimer's disease patient. In order to determine the informativeness of the signals, an analysis of variance (ANOVA) was conducted with a significance level ($p$) of 0.05. A maximum F-ratio ($F_m$) value was used to select the most informative signal type.

Figure 14 shows the $F_m$ values for all types of signals. Among all types of filtered signals, the approximation coefficient A3 (0–32.5 Hz) provides the most information. Additionally, the coefficients A1, A2 and A4, as well as XFEEG(t) have relatively high $F_m$ values. In terms of detail factor analysis, D4 (16.25–32.5 Hz) is the most informative. The coefficients A6, D1 and D2 do not allow us to state that the EEG signals from healthy and sick patients differ using SVDEn, as they are greater than the significance level ($p = 0.05$).

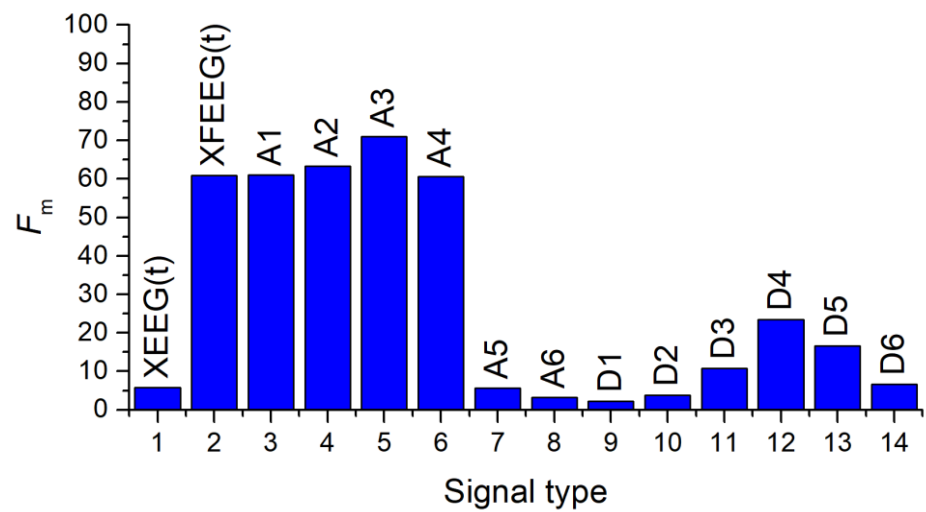

**Figure 14.** Distribution of $F_m$ values for various types of signals calculated using the SVDEn feature. $F_m$—maximum F-ratio values for all 19 channels. The figure reveals the most informative types of signals.

3.4.2. Influence of the Entropy Calculation Method on the Separation of EEG Signals

In order to determine the optimal method for calculating entropy using A3 coefficients, the following entropies were tested: SVDEn, PermEn, SampEn, CoSiEn, FuzzyEn, DistEn, PhaseEn, BubbleEn, AttnEn, IncEn, NNetEn. The $F_m$ values for various entropies are shown in Figure 15. In order to maximize $F_m$, the entropy parameters were chosen in a manner that maximized the value of $F_m$. In this case, FuzzyEn $F_m = 140$ seems to produce the best results. It is found that all three methods (SampEn, CoSiEn, FuzzyEn) yield a relatively high result when vectors that are similar to the built-in matrix are taken into consideration. The worst result comes from DistEn—$F_m = 6.8$. NNetEn shows that there is a sufficiently good separation of the signals from healthy and sick patients at $F_m = 40.29$.

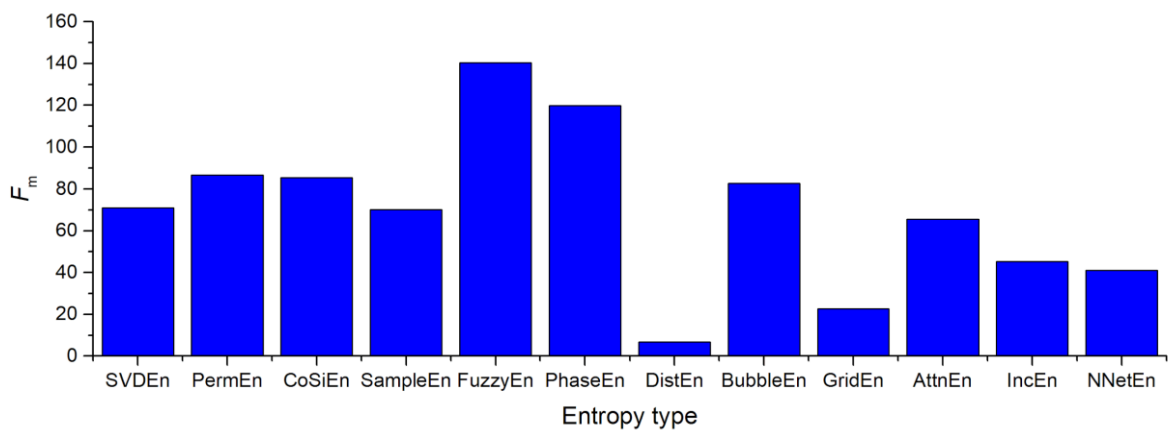

**Figure 15.** Distribution of the $F_m$ value when using the A3 coefficients for various entropy calculation methods. The figure shows the ratio of entropy efficiencies.

In Figure 16, the best F-ratio values for different entropy methods are shown for each channel. It was found that channel number 10 (Cz) was the most significant among the channels processed. Additionally, good separation can be observed in channels 7 (F8), 9 (C3), 15 (Pz), and 16 (P4). On channels 5, 11, 12, 13, and 18, NNetEn ranks among the leaders in terms of feature strength, and on channel 11, it shows the best results.

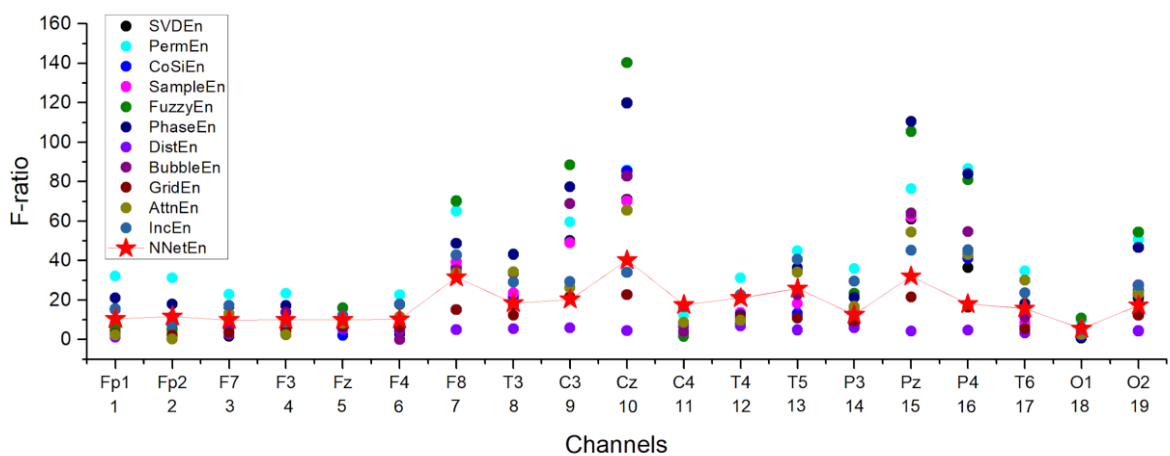

**Figure 16.** Distribution of the F-ratio value for various entropy calculation methods using A3 coefficients. The figure shows the performance of NNetEn as a feature for different channels.

In Figure 17, F-ratio distributions for Dataset 1 (Figure 17a) and Dataset 2 (Figure 17b) are shown based on the settings number and channel. The best reservoir filling methods for Dataset 1 are M3, M4, and M5, and for Dataset 2 M1, M3, and M4. In Dataset 1, channels 7, 8, 10, 11, 12, and 15 provide the best signal recognition, while in Dataset 2, channels 7, 9, 10, 13, 15, and 16 provide the best signal recognition.

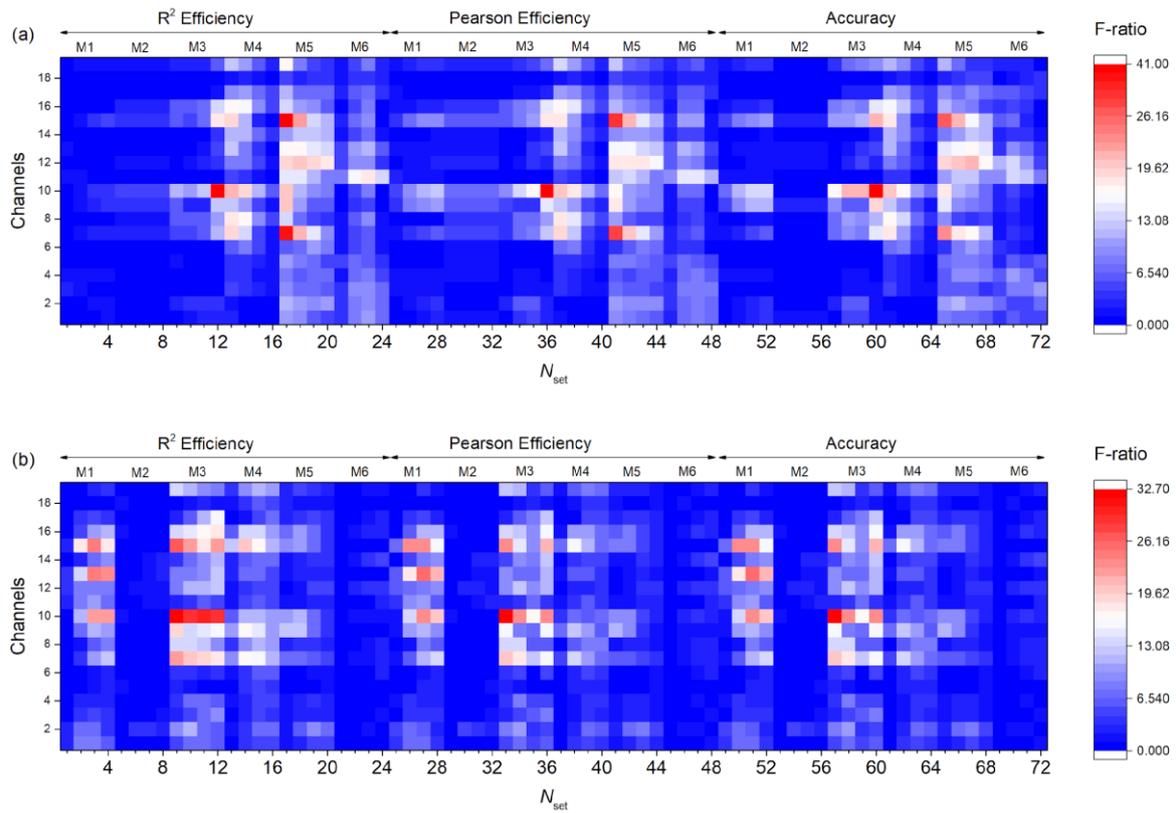

**Figure 17.** F-ratio distribution depending on the setting number and channel for Dataset 1 (**a**) and Dataset 2 (**b**). The most effective settings are in red.

In terms of the single feature, NNetEn generally performed quite well in comparison with other entropies. The synergistic effect of standard entropies with NNetEn(D1,1, $N_{set}$ = 60) is shown in Figure 18. A synergistic effect was observed when almost all entropies were paired with NNetEn, increasing the total $A_{RKF}$ value (Figure 18a). The dependence of $K_{syn}$ on pair type can be seen in Figure 18b. The most effective synergistic effects were observed for IncEn, AttnEn, SampEn, and SVDEn.

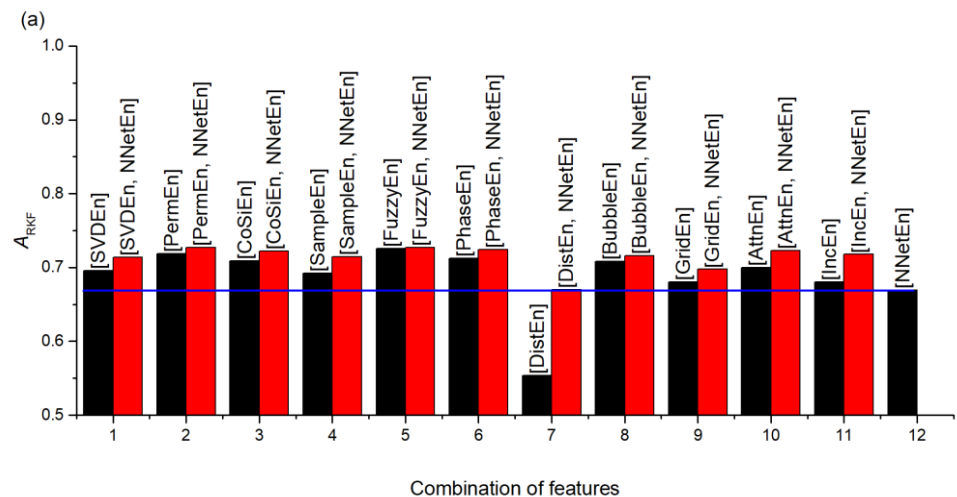

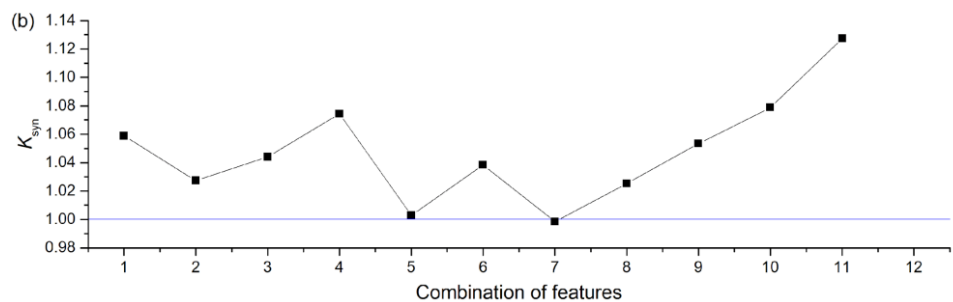

**Figure 18.** Synergistic effect of standard entropies with NNetEn(D1,1, *Nset* = 60) shown based on accuracy $A_{RKF}$ (**a**), and coefficient $K_{syn}$ (**b**). The figure demonstrates an increase in the efficiency of classification, when using NNetEn as a paired feature to other entropies.

*3.5. Features of the Python Execution of the NNetEn Algorithm*

Python programming language is an interpreted language, and calculations take a long time compared with pre-compiled programs like Delphi. A just-in-time (JIT) compiler was used in Numba's open-source project in order to speed up Python source code execution. Numba decorated functions are compiled into machine code "just-in-time" for execution, enabling your code to run at native machine code speed after they are called. It was possible to reduce the time required for Python to calculate entropy by using Numba. A summary of the average network training stage (Stage 9) for NNetEn(D1, 1, Ep5, M2, Acc) is shown in Table 2.

**Table 2.** Average durations of the network training stage (Stage 9) using different programming languages.

|  | $\mu = 1$ | $\mu = 1$ |
| --- | --- | --- |
|  |  |  |
|  |  |  |
|  |  |  |

Table 2 shows that the calculation of NNetEn entropy without Numba takes 11.5 s, while the calculation time with the Numba compiler is 3.7 s. In fact, this execution speed is even faster than the Delphi program, indicating this is an optimal execution speed. The Numba compiler may be overspeeding calculations due to the use of vector operations and caching.

**4. Conclusions**

Feature extraction is a key part in time series analyzing and classification procedures. Entropy measures are well-known time series features that are widely described in the literature. Traditional entropy measures are not appropriate for short and noisy time series. Moreover, they are sensitive to initial parameters that leads to inaccurate results. NNetEn overcomes these difficulties, and it has been applied to solving practical problems successfully. However, there were no functional modules for calculating NNetEn in common programming languages. To overcome this shortcoming, we apply Python language to implement the NNetEn algorithm. To increase the speed of calculations, the SARS-CoV-2-RBV1 dataset is used as the input data instead MNIST-10 dataset. Based on the NNetEn algorithm, three neural network entropy metrics (NNetEn, R2 Efficiency and Pearson Efficiency) are provided. The sine chaotic map, ANOVA and F-score are used for investigating the efficiency of new metrics in distinguishing between different signals. The results demonstrate that the new metrics are an efficient tool for distinguishing between the time series. The EEG classification of patients with Alzheimer's disease (36 people) and those in the control group (29 healthy people) is illustrated as a practical application. The SVM algorithm is used for the classification. For some channels, NNetEn ranks among the leaders in terms of feature strength; and on channel 11, it demonstrates the best results. Our computations confirm that a combination of classical entropy measures and NNetEn method provides results that are more accurate. Our computations demonstrate the synergistic effect of increasing classification accuracy when applying traditional entropy measures and the NNetEn concept conjointly.

In practice, it is important to understand what entropies to use for signal separation and classification. However, each case requires its own parameters and types of entropies.

NNetEn is the first entropy that does not use probability distribution function and provides useful information about the signal in the classification problems. For each individual task, a specific entropy measure must be tested, as it is difficult to predict how effective the measure will be in the given settings. Based on the examples presented, entropy depends in a complex way on the calculation parameters, and it is possible to extract useful information from a change in entropy. It is common for signals with similar dynamics to have different entropy correlations based on their parameters. Implementing NNetEn in Python will allow the scientific community to apply the algorithm to identify a class of problems that are effectively solved by NNetEn. The review of previous studies demonstrates the NNetEn's success in practical problems. Moreover, NNetEn is illustrated with a practical example. In terms of feature strength, NNetEn ranks among the leaders for the classification of EEG signals from certain channels.

The challenge of filling the reservoir matrix is the availability of many options. In Section 2.3, we have introduced six basic filling methods that affect the obtained results. Therefore, the method is considered as NNetEn input parameter. [ 1 shows the general concept of NNetEn calculation without specifying the method, in contrast to the algorithm defined in Figure 2. Therefore, the search for new types of reservoirs in the concept of NNetEn calculation can be a subject of further research. However, it is already possible to use the matrix reservoir model with six filling methods in practical signal classification cases.

Parallelizing the NNetEn algorithm, applying the new metric on different practical problems and investigating the effects on new metrics on different classification algorithms can be considered as future research directions.

Although entropy is a measure of chaos, results demonstrate that two signals for different entropy parameters can have a different ratio. Therefore, the ability of entropy to sense subtle differences in signals without tying it to the concept of chaos is a feature of entropy functions. It can be apparent reason for many different methods and algorithms for calculating entropy, including a new algorithm presented in this study.

**Supplementary Materials:** Python package for NNetEn calculation involved in this study is publicly available on GitHub. https://github.com/izotov93/NNetEn (accessed on 26 April 2023) . Examples of NNetEn calculation (examples.zip).

**Author Contributions:** Conceptualization, A.V., M.B., Y.I., M.M. and H.H.; methodology, A.V. and M.B.; software, A.V., M.B. and Y.I.; validation, M.B.; formal analysis, M.M. and H.H.; investigation, A.V. and M.B.; resources, A.V.; data curation, A.V.; writing—original draft preparation, A.V., M.B., Y.I., M.M. and H.H.; writing—review and editing, A.V., M.B., Y.I., M.M. and H.H.; visualization, A.V., M.B. and Y.I.; supervision, A.V.; project administration, A.V.; funding acquisition, A.V. All authors have read and agreed to the published version of the manuscript.

**Funding:** This research was supported by the Russian Science Foundation (grant no. 22-11-00055, https://rscf.ru/en/project/22-11-00055/, accessed on 30 March 2023).

**Data Availability Statement:** The data used in this study can be shared with the parties, provided that the article is cited.

**Acknowledgments:** The authors express their gratitude to Andrei Rikkiev for valuable comments made in the course of the article's translation and revision. Special thanks to the editors of the journal and to the anonymous reviewers for their constructive criticism and improvement suggestions.

**Conflicts of Interest:** The authors declare no conflict of interest.